%% file: root.tex
\documentclass[letterpaper, 10 pt, conference]{ieeeconf}  % Comment this line out if you need a4paper

\IEEEoverridecommandlockouts                              % This command is only needed if 
\newcommand{\etal}{\textit{et al.}}
\usepackage{graphicx} 
\usepackage{amsmath}
\usepackage[pagebackref,breaklinks]{hyperref}
\hypersetup{colorlinks,linkcolor={red},urlcolor={black}} 
\usepackage[capitalize]{cleveref}
\usepackage{amsmath}
\usepackage{graphicx}
\usepackage{booktabs}
\usepackage{multirow}
\usepackage{array}
\usepackage{tikz}
\usepackage{amssymb}
\usepackage{pdfrender}
\usepackage{pifont}
\usepackage{bbding}
\usepackage{makecell}
\title{\LARGE \bf
Unified Human Localization and Trajectory Prediction\\ with Monocular Vision
}

\author{Po-Chien Luan$^{1*}$, Yang Gao$^{1*}$, Céline Demonsant$^{1}$ and Alexandre Alahi$^{1}$% <-this % stops a space
\thanks{$^{*}$Equal contribution}
\thanks{$^{1}$EPFL, Lausanne, Switzerland; Email: firstname.lastname@epfl.ch
       }%
\thanks{Corresponding author: Po-Chien Luan, po-chien.luan@epfl.ch}
}

\begin{document}

\maketitle
\thispagestyle{empty}
\pagestyle{empty}

%%%%%%%%%%%%%%%%%%%%%%%%%%%%%%%%%%%%%%%%%%%%%%%%%%%%%%%%%%%%%%%%%%%%%%%%%%%%%%%%
\begin{abstract}

\input{sections/abstract}

\end{abstract}

%%%%%%%%%%%%%%%%%%%%%%%%%%%%%%%%%%%%%%%%%%%%%%%%%%%%%%%%%%%%%%%%%%%%%%%%%%%%%%%%
\section{INTRODUCTION}

\input{sections/intro}

\section{RELATED WORK}
\input{sections/related}

\section{METHOD}
\input{sections/method}

\section{EXPERIMENTS}
\input{sections/experiments}

\section{CONCLUSIONS}

\input{sections/conclusions}


%%%%%%%%%%%%%%%%%%%%%%%%%%%%%%%%%%%%%%%%%%%%%%%%%%%%%%%%%%%%%%%%%%%%%%%%%%%%%%%%

%%%%%%%%%%%%%%%%%%%%%%%%%%%%%%%%%%%%%%%%%%%%%%%%%%%%%%%%%%%%%%%%%%%%%%%%%%%%%%%%

%%%%%%%%%%%%%%%%%%%%%%%%%%%%%%%%%%%%%%%%%%%%%%%%%%%%%%%%%%%%%%%%%%%%%%%%%%%%%%%%

% \section*{ACKNOWLEDGMENT}

% ...
% \section*{REFERENCES}

\newpage
\bibliographystyle{IEEEtran}
\bibliography{root}  % .bib

\end{document}

%% file: sections/abstract.tex
Conventional human trajectory prediction models rely on clean curated data, requiring specialized equipment or manual labeling, which is often impractical for robotic applications. The existing predictors tend to overfit to clean observation affecting their robustness when used with noisy inputs. In this work, we propose MonoTransmotion (MT), a Transformer-based framework that uses only a monocular camera to jointly solve localization and prediction tasks. Our framework has two main modules: Bird's Eye View (BEV) localization and trajectory prediction. The BEV localization module estimates the position of a person using 2D human poses, enhanced by a novel directional loss for smoother sequential localizations. The trajectory prediction module predicts future motion from these estimates. We show that by jointly training both tasks with our unified framework, our method is more robust in real-world scenarios made of noisy inputs. We validate our MT network on both curated and non-curated datasets. On the curated dataset, MT achieves around 12\% improvement over baseline models on BEV localization and trajectory prediction. On real-world non-curated dataset, experimental results indicate that MT maintains similar performance levels, highlighting its robustness and generalization capability. The code is available at \href{https://github.com/vita-epfl/MonoTransmotion}{\color{magenta}https://github.com/vita-epfl/MonoTransmotion}.

%% file: sections/intro.tex
Predicting human motion is crucial for robots navigating crowded environments. Generally, prediction relies on perception. The perception module detects and localizes surrounding agents, while the prediction module forecasts the future trajectories of these agents. High-quality perception and prediction are essential for preventing potential collisions, ensuring safe human-robot interactions, and improving robotic planning.

\par
\begin{figure}[!tbp]
\centering
    \includegraphics[width=1.0\linewidth]{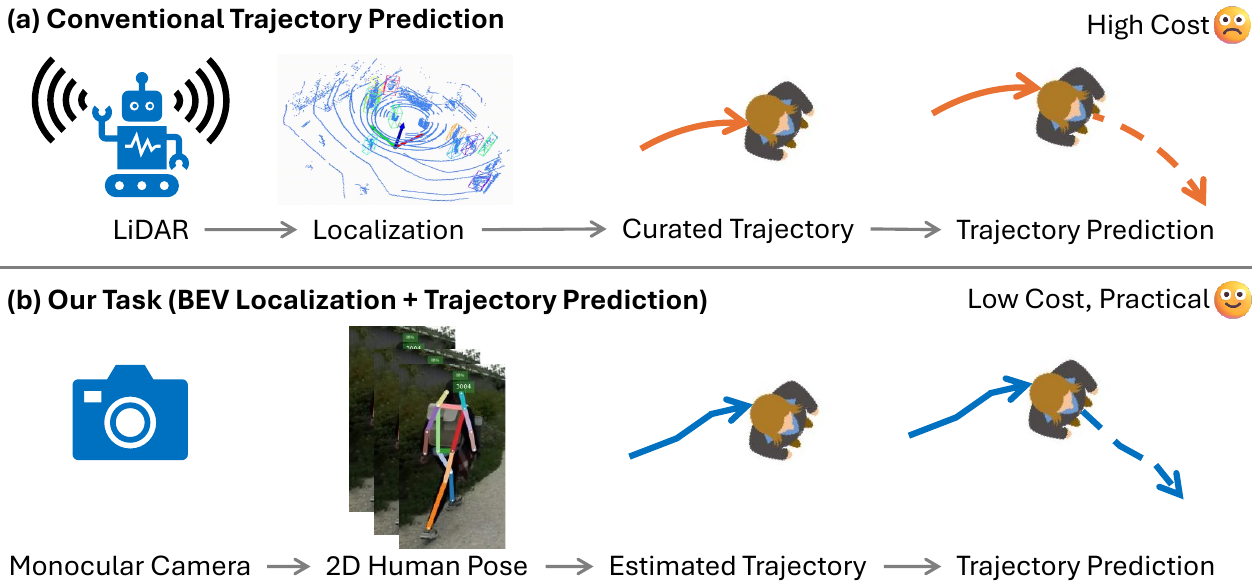}
    \caption{\textbf{Comparison of our task and conventional pipeline.} (a) Conventional settings directly obtain curated localization from LiDAR or other sensors \cite{paez20213d}. (b) Our task focuses on leveraging only a monocular camera to estimate BEV localization and using the estimates to predict trajectories. We present a unified model that jointly solves localization and prediction.} 
    \vspace{-.6cm}
    \label{fig:pull}
\end{figure}

As illustrated in \Cref{fig:pull}, conventional methods \cite{alahi2016social,saadatnejad2024socialtransmotion} rely on curated data from sources like LiDAR or fixed cameras, often combined with some post-processing to predict future trajectories. These methods typically assume that the perception module provides ground-truth trajectories.  However, relying on curated data is often impractical for robotics. In real-world scenarios, the trajectory prediction module receives data from upstream processes, including object detection and localization. Here, localization refers specifically to estimating the positions of humans in the BEV using a monocular camera, which differs from the traditional notion of robot localization in SLAM contexts. Localization is particularly critical \cite{xu2024towards,wang2021fcos3d,bertoni2019monoloco}. Noisy data from these sources can significantly degrade the performance of trajectory predictions.

\par

LiDAR systems offer precise depth information for localization, but their high cost limits their practicality compared to more affordable camera-based solutions. Fixed BEV cameras \cite{pellegrini2009you} can directly yield trajectories but are not helpful for mobile robots. Our goal is to develop a model that operates without costly hardware and is suitable for mobile robots. To achieve this, we shift our attention from LiDAR-based approaches to camera-based localization, considering the movement of the ego robot and aiming to create a cost-effective yet robust solution.

% \end{itemize}

\par

We present MonoTransmotion (MT), a transformer-based model that integrates both localization and trajectory prediction modules in a unified architecture. 
The human localization module estimates the position of a person using 2D human poses, enhanced by a novel directional loss for smoother sequential localizations. The trajectory prediction module predicts future motion from these estimates. We show that by jointly training both task with our unified framework, our method is more robust in real-world scenarios made of noisy inputs. We evaluate our MT model on two datasets: an autonomous driving dataset (NuScenes~\cite{caesar2020nuscenes}), and Head-Mounted Egocentric Dataset for Trajectory Prediction in Blind Assistance Systems (HEADS-UP~\cite{VIP2024}). MT effectively estimates BEV localization and accurately predicts trajectories. Here are the contributions summarized into threefold:

\begin{itemize}
    \item We propose MonoTransmotion (MT), a Transformer-based framework that uses only monocular cameras for joint human localization and trajectory prediction, eliminating the need for expensive sensors like LiDAR or curated datasets, making it more accessible for real-world applications such as mobile robotics and visually impaired assistance.
    
    \item We introduced a directional loss function to improve the smoothness and accuracy of human localization. This ensures that MT can handle dynamic, real-world environments with greater precision compared to conventional methods.
    
    \item We validate MT on both curated and non-curated datasets, demonstrating strong performance even in noisy, real-world conditions, such as using a head-mounted camera for visually impaired assistance. This highlights MT's robustness and adaptability for practical, real-time deployment in everyday scenarios.    
\end{itemize}

%% file: sections/related.tex
%low-cost
%trajectory prediction relies on GT from LiDar. Previous works use curated data
%Real-time for robotics
%More practical settings: how difficult to get curated data from Lidar
\subsection{Monocular localization}
% our highlights
% 1. fast! use 2d keypoints instead of raw image
% 2. better performance! use sequential poses instead of single-frame pose

Monocular localization, which involves extracting the location of objects from single-view images, has gained significant attention recently due to its cost-effectiveness. Most related works \cite{wang2021fcos3d, xiong2023you, zhang2024decoupled, liu2023monocular, wu2023monopgc} treat localization as part of the broader 3D object detection task. While LiDAR data can enhance performance in 3D object detection \cite{chen2021monorun, reading2021categorical}, its high cost limits its accessibility. Brazil~\etal \cite{brazil2020kinematic} demonstrated the advantages of incorporating temporal information into localization. For human objects, Monoloco \cite{bertoni2019monoloco} introduced a lightweight model that performs monocular localization using 2D pose keypoints rather than raw images, marking a significant step forward. Building on these foundational works, we propose a Transformer-based sequential Monoloco, which leverages temporal information to further enhance monocular localization performance.

\subsection{Human trajectory prediction}
% our highlights
% 1. no need for curated data (GT BEV coordinates from Lidar)
% 2. use monocular image as input

Human trajectory prediction has emerged as a vital research area in recent years, with deep learning techniques significantly advancing data-driven methods \cite{alahi2016social, gupta2018social, girgis2022latent}, surpassing traditional hand-crafted approaches \cite{helbing1995social, fujii2013extended}. Various model architectures have proven effective for trajectory prediction. Long Short-Term Memory (LSTM) networks \cite{alahi2016social, kothari2021human} excel in modeling social interactions through pooling mechanisms, while Convolutional Neural Networks (CNNs) \cite{nikhil2018convolutional} enhance parallel processing and temporal representation. Generative Adversarial Networks (GANs) \cite{gupta2018social, sadeghian2019sophie} and Conditional Variational Autoencoders (CVAEs) \cite{schmerling2018multimodal, ivanovic2020multimodal} are widely used to generate multiple potential trajectories in stochastic settings. Recently, diffusion-based models \cite{wang2024optimizing} have explored controllable trajectory generation, and attention-based Transformer networks \cite{saadatnejad2024socialtransmotion, girgis2022latent} have gained attention for their advantages in multi-modal learning \cite{bachmann2022multimae} and data scaling \cite{feng2024unitraj}.
However, a common limitation among current human trajectory predictors is their reliance on curated data, such as LiDAR-derived BEV coordinates, which restricts their practical application. This has motivated us to propose a unified model that directly takes monocular images as input and predicts future BEV coordinates as output, eliminating the need for expensive and curated data.

\subsection{End-to-end trajectory prediction}
% our highlights
% 1. end-to-end for human
% 2. one monocular camera instead of multiple camera, Lidar, ...

Recently, there has been increasing interest in end-to-end trajectory prediction, as joint learning can help reduce accumulated errors \cite{hu2023planning}. Several pioneering studies \cite{luo2018fast, liang2020pnpnet, philion2020lift, hu2021fiery, gu2023vip3d} have explored end-to-end systems in vehicle motion prediction and planning. Most of these works \cite{luo2018fast, liang2020pnpnet, philion2020lift, gu2023vip3d} utilize LiDAR to collect curated data on neighboring agents, enabling downstream modules to leverage precise localization information.
One closely related work, FIERY \cite{hu2021fiery}, directly predicts future BEV vehicle trajectories using multiple surrounding monocular cameras. However, predicting human motion significantly differs from predicting vehicle motion due to the non-rigid nature of the human body, which can move freely in environments without defined lanes. To bridge this gap and extend end-to-end systems to human motion, we aim to comprehensively investigate the potential of predicting BEV human trajectories using only a single monocular camera.

%% file: sections/method.tex
\begin{figure}[!tbp]
    \centering
    \includegraphics[width=\linewidth]{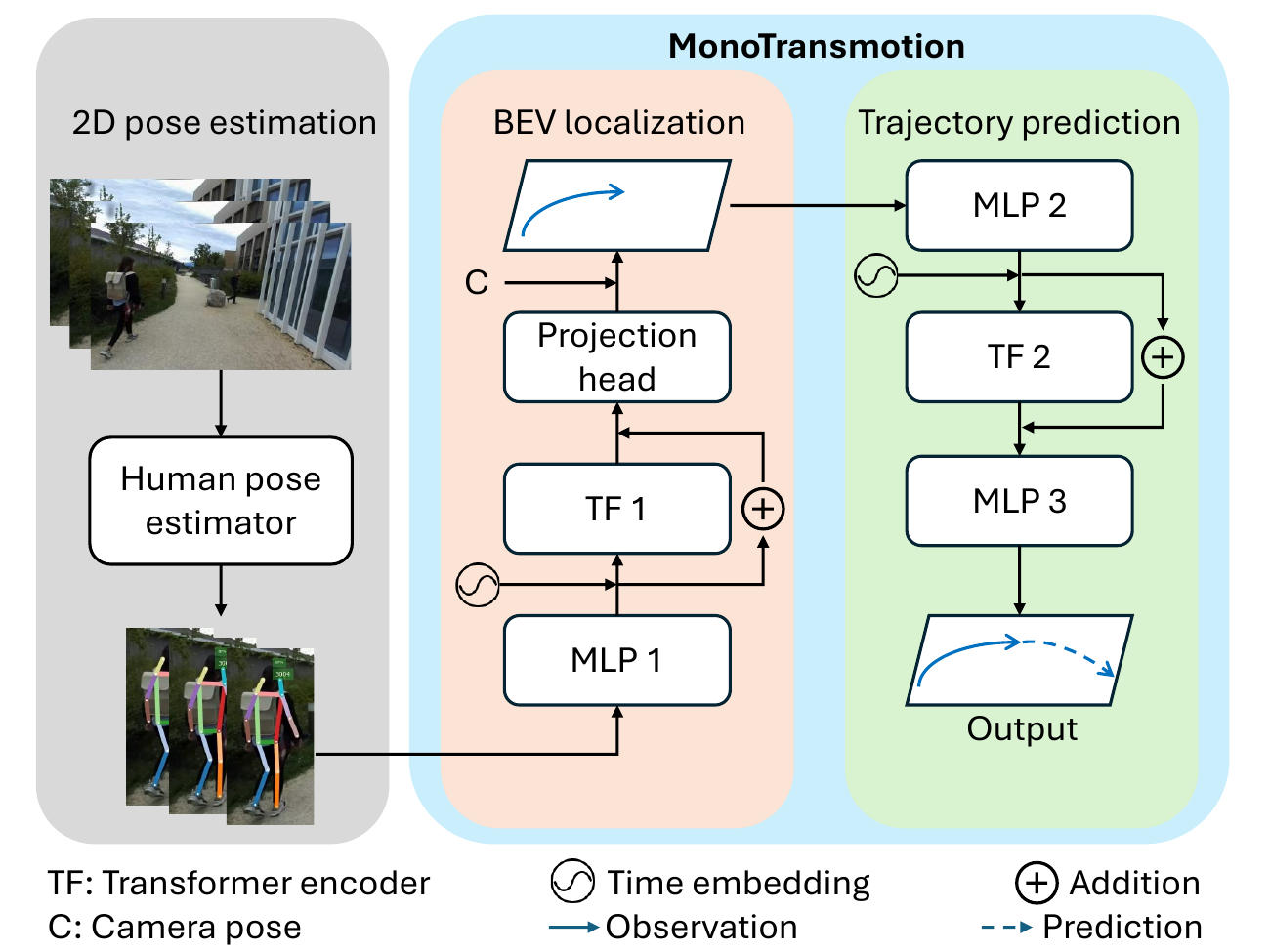}
    \caption{\textbf{The overview of the complete pipeline}. MonoTransmotion (MT) utilizes estimated 2D human poses as input to simultaneously address both BEV localization and trajectory prediction tasks.}
    \vspace{-.8cm}
    \label{fig:overview}
\end{figure}
MT integrates BEV localization and trajectory prediction modules, utilizing sequential 2D human poses as input instead of images to enhance efficiency. An overview of the complete pipeline is shown in~\Cref{fig:overview}. In Section~\ref{sec:pf}, we describe the inputs and outputs of our task, and in Section~\ref{sec:MT}, we detail each module of the MT framework.

\subsection{Problem formulation}
\label{sec:pf} 
We define the observation time steps as $t=1,2,...,T_{obs}$, and the prediction time steps as $t=T_{obs+1},T_{obs+2},...,T_{pred}$. Let the sequence of 2D human poses be denoted as $X^{2dP} = [X_{1}^{2dP}, X_{2}^{2dP}, ..., X_{T_{obs}}^{2dP} ]$. This sequence is used to estimate the BEV trajectory, $X^{Traj} = [X_{1}^{Traj}, X_{2}^{Traj}, ..., X_{T_{obs}}^{Traj} ]$. Finally, the future BEV trajectory is predicted as $Y^{Traj} = [Y_{T_{obs+1}}^{Traj}, Y_{T_{obs+2}}^{Traj}, ..., Y_{T_{pred}}^{Traj} ]$.

\subsection{MonoTransmotion}
\label{sec:MT}

MT uses preprocessed 2D human pose keypoints as input to perform BEV localization followed by trajectory prediction. Below, we detail the preprocessing, BEV localization, and trajectory prediction processes.
\subsubsection{Preprocessing}
The preprocessing stage involves detecting, tracking, and estimating 2D human poses. First, we detect pedestrians and estimate their 2D poses. To assign IDs to pedestrians, we consider two different conditions: for the curated dataset, we directly use ground-truth labels, following the conventional trajectory prediction pipeline; for the non-curated dataset, we employ serial algorithms ~\cite{kreiss2021openpifpaf,maggiolino2023deep} to fit real-world settings. It is important to note that our primary focus is on understanding how monocular BEV localization impacts trajectory prediction. Therefore, we utilize these off-the-shelf models to concentrate on the BEV localization and trajectory prediction modules, which are the core areas of our research.

\subsubsection{BEV localization} To improve localization accuracy and smoothness, we incorporate sequential information and a well-designed loss function. The structure of the BEV localization module is illustrated in~\Cref{fig:overview}.
We input sequential 2D human poses $X^{2dP}=[X_{1}^{2dP},...,X_{T}^{2dP}]$ into the model. First, we use Multi-Layer Perceptron (MLP) layers to get the embeddings. To incorporate temporal information, we add learnable positional embeddings to these embeddings. The enhanced embeddings are then passed through a Transformer encoder. To project the output embeddings into BEV localization, each time step is connected to MLP layers. The outputs are the local coordinates from the camera view, denoted as $X^{local}=[X_{1}^{local},..., X_{T}^{local}]$. Specifically, we represent these local coordinates in spherical coordinates as $X_{t}^{local}=[r_{t},\theta_t,\phi_t]$. To form a complete trajectory in BEV, we project the local coordinates into global coordinates, denoted as $X^{Traj}=[X_{1}^{Traj},..., X_{T}^{Traj}]$.
\par
\noindent\textbf{Localization loss.} Our goal is to estimate high-quality trajectories of the observation to predict the future. Apart from the precision of localization, direction is also critical. Most trajectory prediction models utilize velocity so that wrong directions would mislead the future trajectories \cite{korbmacher2022review}. We can design localization based on two parts: precision and direction. To obtain precise localization, we follow the loss of Monoloco \cite{bertoni2019monoloco}, Laplace loss $L_{Laplace}$, and L1 loss $L_{l1}$. These two loss functions regularize the position error. Unlike conventional BEV localization models, our task is to localize a trajectory sequentially rather than in a single position. Monoloco loss only reflects the precision of the localization. To regularize the direction, we need to consider all localizations as an entity. The velocity of each time step is important to trajectory prediction. Inaccurate velocity might mislead the model predicting trajectory in the wrong direction. As a result, we propose a directional loss $L_{dir}$ to regularize the direction of the estimated trajectories. We can summarize the localization loss as follows:
\begin{equation}
L_\text{loc}=L_{Laplace}+L_{l1}+\lambda_{dir}L_{dir},  
\end{equation}
where $\lambda_{dir}$ is a weight to control the importance of regularizing direction. $L_{Laplace}$ and $L_{l1}$ can be expressed as follows:
\begin{equation}
L_{Laplace}= \sum_{t=1}^{T} \frac{|1-r_t/r^*_t|}{b_t}+\log(2b_t),
\end{equation}
\begin{equation}
    L_{l1}= \sum_{t=1}^{T} |\theta_t - \theta^*_t|+|\phi_t - \phi^*_t|,
\end{equation}
where notations with a star represent ground truth. $b$ is the standard deviation of predicted depth. $L_{Laplace}$ indicates that we use $r$ and $b$ to represent aleatoric uncertainty \cite{bertoni2019monoloco}. $L_{l1}$ supervises the angles of spherical coordinates. To consider the characteristics of trajectories, we propose $L_{dir}$ as follows:
\begin{equation}
    L_{dir}= \frac{-1}{T-2}\sum_{t=2}^{T-1}\frac{v_t}{\|v_t\|}\cdot \frac{v^*_t}{\|v^*_t\|},
\end{equation}
where $v_t=x^{Traj}_t-x^{Traj}_{t-1}$ is the velocity of the trajectory at time $t$. $L_{dir}$ regularizes the cosine angle between the estimated and ground-truth trajectory. 
\subsubsection{Trajectory prediction}
The trajectory prediction module focuses on predicting future BEV global coordinates from the observed ones, as illustrated in \Cref{fig:overview}. Once the global trajectories $X^{Traj}=[X_{1}^{Traj},X_{2}^{Traj},...,X_{T}^{Traj}]$ are obtained, an MLP is used to project these coordinates into a hidden dimension, followed by the addition of time embeddings to encode temporal information. The trajectory tokens are then fed into a Transformer-based trajectory predictor \cite{saadatnejad2024socialtransmotion}, which learns motion representations. The final output is the predicted BEV global trajectories over the prediction horizon, denoted as $Y^{Traj} = [Y_{T_{obs+1}}^{Traj}, Y_{T_{obs+2}}^{Traj}, ..., Y_{T_{pred}}^{Traj} ]$. 

\par
\noindent\textbf{Trajectory prediction loss.} We use the MSE Loss for the trajectory prediction task.
\begin{equation}
    L_{Traj}=\frac{1}{(T_{\text{pred}} - T_{\text{obs}})} \sum_{t=T_{obs+1}}^{T_{pred}}\Vert Y_{t}^{Traj}-\hat{Y}_{t}^{Traj}\Vert_2,
\end{equation}
where $\hat{Y}_{t}^{Traj}$ is the estimated location of prediction horizon at time step $t$.

\subsubsection{Training and Inference} 

During training, we input the entire sequence of keypoints into the model, setting $T=T_{pred}$. The BEV localization module estimates the full trajectory, including both the observation and prediction phases. We supervise trajectory prediction using estimated trajectories instead of ground truth to maintain trajectory continuity, as ground truth can introduce abrupt changes between $T_{obs}$ and $T_{obs+1}$, potentially degrading prediction performance. During inference, MT observes keypoints over $T_{obs}$ time steps, and the BEV localization module estimates the trajectory based on these observations. The trajectory prediction module then forecasts the future trajectory over the remaining $T_{pred}-T_{obs}$ time steps.

%% file: sections/experiments.tex
We conducted a series of experiments to comprehensively evaluate the performance of MT. First, we assess MT qualitatively and quantitatively on BEV localization and trajectory prediction tasks. Second, we discuss the importance of modeling BEV localization as a time series problem. Additionally, we explore various training strategies to optimize MT’s performance. Finally, we demonstrate MT on a real-world, non-curated dataset designed for visually impaired individuals, showcasing its ability to handle noisy data effectively.

\par

\noindent\textbf{Curated dataset.} Our goal is to implement MT in real-world settings. To prepare for this, we first evaluate MT on the NuScenes dataset \cite{caesar2020nuscenes}, which is a comprehensive multimodal dataset widely used in autonomous driving research. Since our task requires 2D keypoints as inputs, we adapt and filter NuScenes to ensure it contains visible keypoints and sufficient sequence length for trajectory prediction. As there is no official benchmark for our specific task, we use the official validation set of NuScenes as our test set.

\par

\noindent\textbf{Non-curated dataset.} One important application of human trajectory prediction is crowd navigation, which is particularly critical for visually impaired people. Human trajectory prediction can help alert users for potential collisions \cite{Liu2022SocialRepresentations, gao2024multi}. We evaluate our approach on the HEADS-UP dataset \cite{VIP2024}, which includes footage from two different campuses, totaling 10,000 examples. One set is used as the training set, while the other serves as the test set.

% One of the applications of human trajectory prediction is crowded navigation, which is critical for visually impaired people. We name this dataset as the visually impaired people dataset (VIP). We design a headband with a camera to assist visually impaired people, shown in Figure \ref{fig:headband}. Human trajectory prediction can alert potential collisions. We collect two pieces of footage from two campuses with a total of 10k examples. One plays as the training set, and the other one is the test set. 
\par
In real-world settings, such as for the HEADS-UP dataset \cite{VIP2024}, there are no human labelers to annotate and refine the data. Therefore, we implement the following pipeline: first, a bottom-up 2D human pose estimation method is used to detect pedestrians and estimate their keypoints. Next, we track and identify pedestrian IDs. For 2D human pose estimation, we use OpenPifpaf~\cite{kreiss2021openpifpaf}, and for tracking and re-identification, we apply Deep OC-SORT~\cite{maggiolino2023deep}. We generate point clouds using the ZED SDK~\cite{Stereolabs} and compute the average 3D positions of the keypoints to extract the pedestrians' BEV coordinates. Additionally, we process depth images using DROID-SLAM~\cite{teed2021droid} to estimate the ego pose.

\par
\noindent\textbf{Metrics.} Our work focuses on BEV localization and trajectory prediction. Instead of using average precision-based metrics, we prioritize evaluating localization error. We adopt the metrics defined by Monoloco~\cite{bertoni2019monoloco} and adapt the average localization error (ALE) to align with the average transition error (ATE) \cite{caesar2020nuscenes}. The key difference is that ALE specifically measures errors in BEV. Since our approach involves localizing entire trajectories, we extend ALE into a sequential version called average displacement error (ADE), which matches the metrics commonly used in trajectory prediction. Specifically, ADE measures the L2 distance between the ground truth and the predicted trajectory. Additionally, we calculate the final displacement error (FDE) for trajectory prediction. All metrics are reported in meters (m).

\par
\noindent\textbf{Baselines.} For BEV localization, we use FCOS3D \cite{wang2021fcos3d} as our image-based baseline and Monoloco~\cite{bertoni2019monoloco} as our keypoints-based baseline. After reviewing recent monocular 3D object detection and localization methods, we found that none were specifically trained on NuScenes, while both Monoloco and FCOS3D are well-trained on this dataset. For trajectory prediction, we use Social-Transmotion (ST)~\cite{saadatnejad2024socialtransmotion}, a state-of-the-art human trajectory prediction model. Additionally, we compare our results to the Kalman Filter (KF)~\cite{kothari2021human}, which serves as the rule-based baseline. We set the observation length to four steps and the prediction length to six steps, synchronizing the frequency of both datasets to 2 Hz.

\subsection{BEV localization and trajectory prediction} 
\label{sec:localization_tp}

\begin{table}[!tbp]
\centering
\begin{tabular}{ccc}
\hline
\hline
      & BEV localization & Trajectory prediction \\ \cline{2-3} 
Model & ADE$\downarrow$              & ADE/FDE$\downarrow$               \\ \hline
FCOS3D~\cite{wang2021fcos3d} + ST~\cite{saadatnejad2024socialtransmotion}     &    1.530              &  2.130/2.438                     \\
Monoloco~\cite{bertoni2019monoloco} + ST~\cite{saadatnejad2024socialtransmotion}   &     1.255            &   1.726/2.169                 \\
MT w/o $L_\text{dir}$(ours)     &  1.202                &    1.632/2.044                   \\
MT (ours)    &       \textbf{1.093}           &     \textbf{1.517/1.931}                 \\ \hline
\hline
\end{tabular}
\caption{Quantitative results on NuScenes dataset~\cite{caesar2020nuscenes}.}
\vspace{-1.cm}
\label{tab:quantative_nuscenes}
\end{table}

\begin{figure*}[!tbp]
    \centering
    \includegraphics[width=0.8\linewidth]{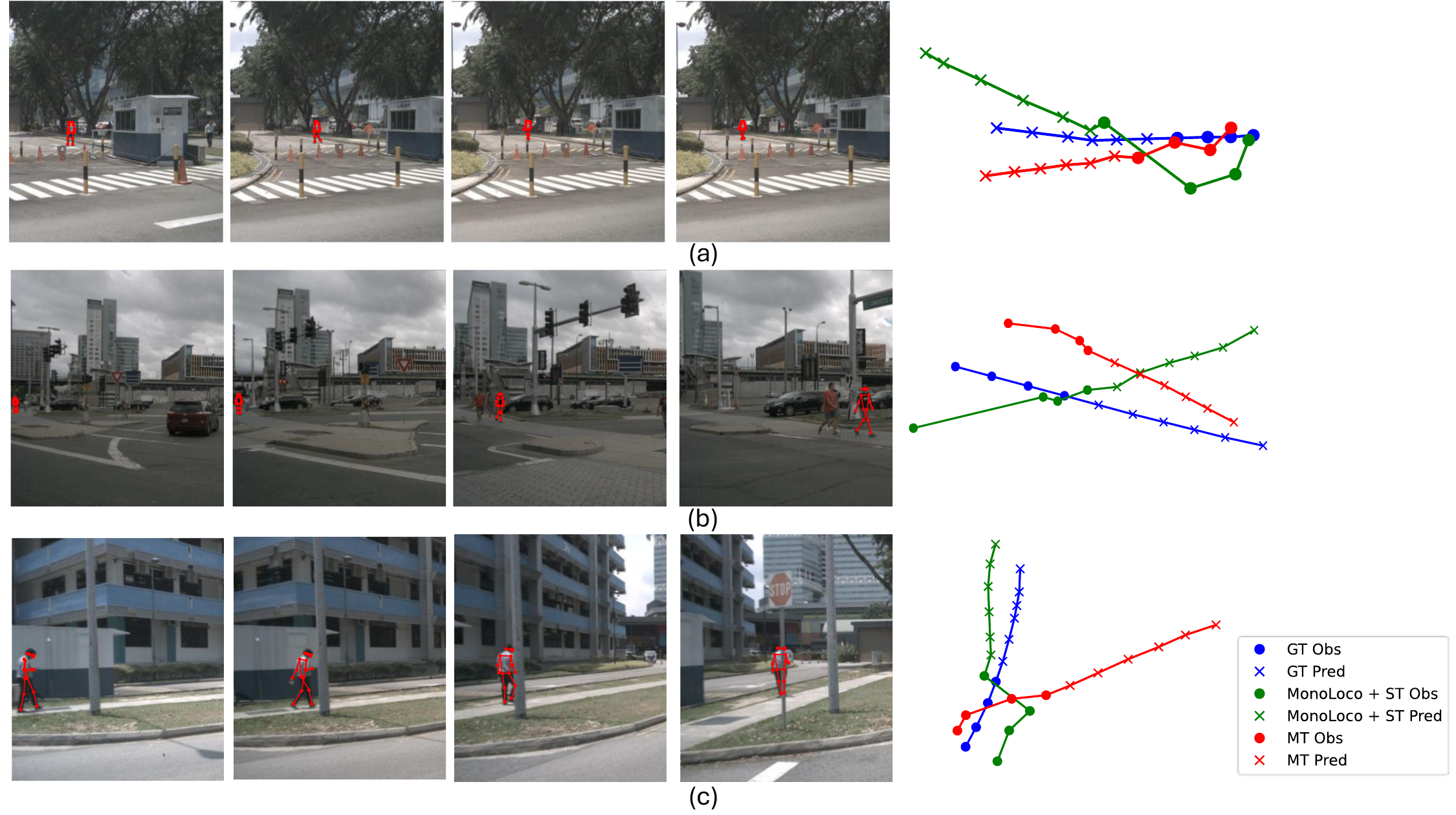}
    \caption{\textbf{Qualitative results on NuScenes.} We visualize four frames of images at consistent intervals, presented in order from left to right. (a) MT achieves better accuracy and smoothness in trajectory prediction. (b) Although there are some deviations in localization, MT maintains the correct direction due to its sequential model and the use of directional loss. (c) MT does not always predict the exact trajectory; however, its sequential approach to localization and trajectory prediction results in a smoother overall trajectory.}
    \vspace{-.5cm}
    \label{fig:quantitative}
\end{figure*}

\Cref{tab:quantative_nuscenes} presents the quantitative results of BEV localization and trajectory prediction, showing that MT outperforms other baselines in both tasks. Keypoints-based methods, such as Monoloco~\cite{bertoni2019monoloco} and our proposed MT, achieve better results in BEV localization compared to the image-based method, FCOS3D~\cite{wang2021fcos3d}. Consequently, we will focus on keypoints-based methods in subsequent experiments. Additionally, the results indicate that improved BEV localization leads to better trajectory prediction. Our proposed directional loss $L_\text{dir}$ further enhances MT’s performance in both BEV localization and trajectory prediction tasks. Qualitative results are illustrated in Figure~\ref{fig:quantitative}, which shows that MT performs lower prediction error and better smoothness. While Monoloco exhibits competitive results with similar inputs, future experiments will primarily compare it with MT.

\begin{figure}[!tbp]
\centering
    \includegraphics[width=0.85\linewidth]{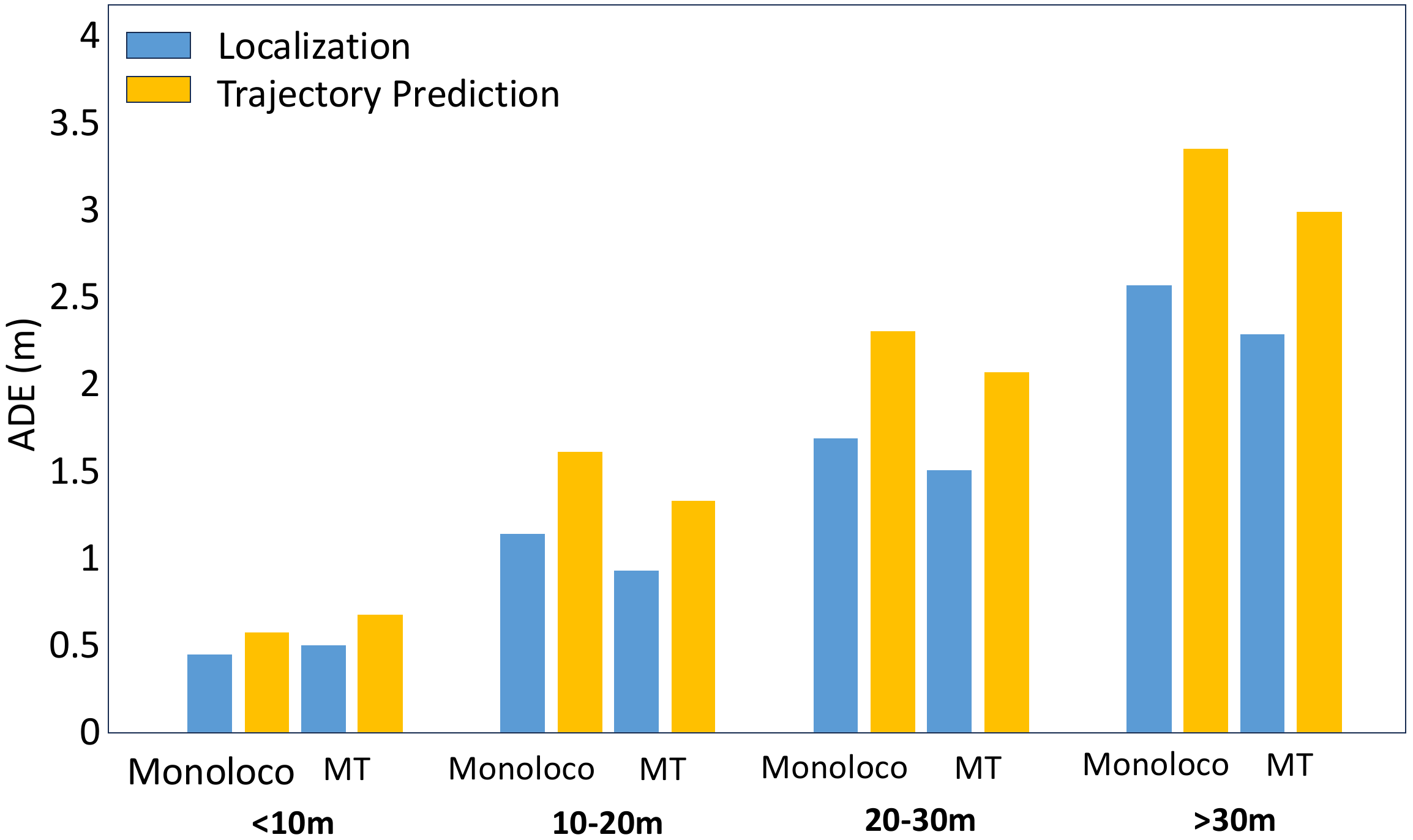}
    \caption{\textbf{Impact of different distances.} We observe that monocular-based BEV localization methods have higher errors when agents are farther away. This increased error in BEV localization significantly impacts the accuracy of trajectory prediction results. }
    \vspace{-.4cm}
    \label{fig:distance}
\end{figure}

\noindent\textbf{Distance to pedestrians}. Monocular 3D localization methods are sensitive to distance; generally, agents that are farther from the ego agent are more ambiguous and challenging for monocular methods to localize accurately. To validate this concept for MT, we categorize the results based on different distance ranges, as shown in \Cref{fig:distance}. The results indicate that both Monoloco~\cite{bertoni2019monoloco} and our MT achieve more precise localization when the targets are closer to the ego agent. We extended this evaluation to trajectory prediction and observed a similar trend. Notably, Monoloco performs slightly better than MT within 10 meters, but it struggles to generalize effectively to longer distances.

\subsection{Ablation study}

\noindent\textbf{Why sequential BEV localization?} We approach BEV localization as a sequential problem using a Transformer-based model and the directional loss $L_\text{dir}$. To explore the impact of this approach, we examine three different structures: 

\begin{enumerate}
    \item Monoloco~\cite{bertoni2019monoloco}: An MLP-based framework.
    \item Transformer-based (TF) Monoloco: This variant uses a Transformer but only takes a single time step of keypoints as input instead of the entire sequence. It encodes seventeen keypoints to estimate localization for each single step.
    \item Our proposed MT: The full sequential model that leverages the power of the Transformer by taking the whole sequence of keypoints as input.
\end{enumerate}

\begin{table}[!tbp]
\centering
\begin{tabular}{cc}
\hline
\hline
Model           & ADE$\downarrow$    \\ \hline
MonoLoco~\cite{bertoni2019monoloco}    & 1.255 \\
TF-MonoLoco & 1.380 \\
MT (ours)   & \textbf{1.087} \\ 
\hline
\hline
\end{tabular}
\caption{Ablation study of different design choices of BEV localization.}
\vspace{-1.cm}
\label{tab:struct}
\end{table}

The comparisons are shown in \Cref{tab:struct}. The results indicate that using single-step keypoints does not fully exploit the capabilities of the Transformer, as encoding only one keypoint set can lead to redundant information and degraded performance. Additionally, non-sequential models are unable to apply loss functions that consider the entire trajectory, such as the proposed $L_\text{dir}$.

\par

\noindent\textbf{Training process.} MT consists of two modules that require training, so exploring different training strategies is essential to achieve optimal results. \Cref{tab:training} lists various training sequences we tested. We found that freezing the BEV localization module yields the lowest localization error, as its performance is already saturated when trained independently from scratch. While joint training improves trajectory prediction, it slightly degrades localization accuracy. Notably, joint training from scratch fails to converge. This issue arises because trajectory prediction heavily depends on the quality of localization. Initially, the BEV localization module produces random results, leading the trajectory prediction module to learn from meaningless data. Consequently, the trajectory prediction loss affects the localization module, resulting in poor performance for both modules. Our primary goal is to achieve accurate trajectory predictions. Therefore, we conclude that training both modules independently until they are well-optimized, followed by a joint fine-tuning process, provides the best results for trajectory prediction.

\begin{table}[!tbp]
\centering
\begin{tabular}{ccc}
\hline
\hline
             & BEV localization & Trajectory prediction \\ \cline{2-3} 
Training order & ADE$\downarrow$  & ADE/FDE$\downarrow$               \\ \hline
\textbf{L} $\rightarrow$ \textbf{T} & \textbf{1.087} & 1.554/1.989 \\
\textbf{L} $\rightarrow$ \textbf{LT}  &  1.119 & 1.542/1.966 \\
\textbf{LT}  &  15.289 & 15.147/15.091  \\
\textbf{L} $\rightarrow$ \textbf{T} $\rightarrow$ \textbf{LT} & 1.093 & \textbf{1.517/1.931} \\ 
\hline
\hline
\end{tabular}
\caption{Ablation study of different training orders. \textbf{L}: training the BEV localization module, \textbf{T}:training the trajectory prediction module, and \textbf{LT}: jointly training both modules.}
\vspace{-1.cm}
\label{tab:training}
\end{table}

\par

\begin{figure*}[!tbp]
    \centering
    \includegraphics[width=0.8\linewidth]{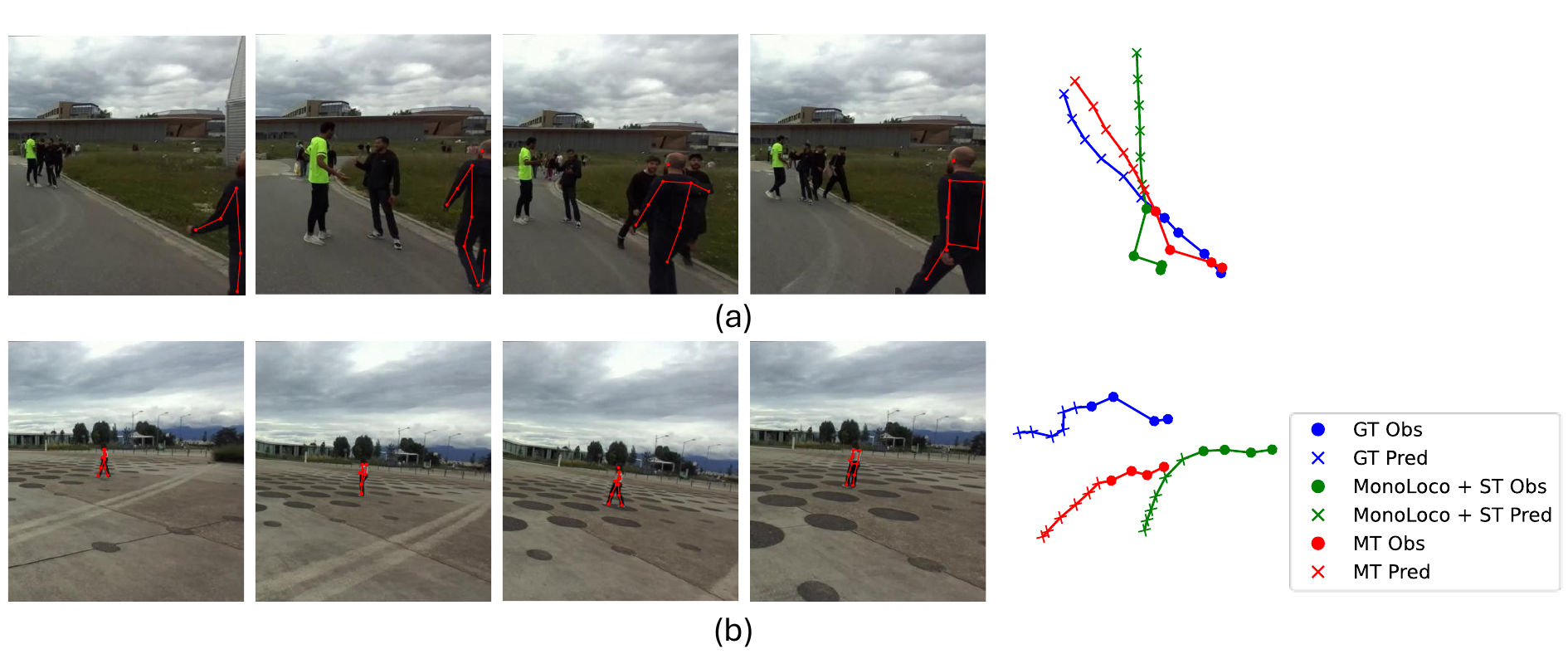}
    \caption{\textbf{Qualitative results on HEADS-UP dataset.} (a) MT offers improved direction and accuracy, even when some keypoints are missing. (b) Monocular methods lose accuracy as the target distance increases. The ground truth from the real-world pipeline contains more noise compared to the carefully curated NuScenes.}
    \vspace{-.3cm}
    \label{fig:quantitative_vip}
\end{figure*}

\noindent\textbf{Handcrafted v.s. learning-based methods.} As discussed in \Cref{sec:localization_tp}, the precision of BEV localization significantly impacts trajectory prediction in learning-based models. To further validate this, we replaced the trajectory prediction module with a Kalman Filter (KF) to evaluate the trajectory prediction results. As \Cref{tab:ablation_module} shows that KF produced consistent outcomes with both ST~\cite{saadatnejad2024socialtransmotion} and MT, demonstrating that improved BEV localization leads to better trajectory prediction, regardless of whether the method is handcrafted or learning-based.

\begin{table}[!tbp]
\centering
\begin{tabular}{ccc}
\hline
\hline
      & BEV localization & Trajectory prediction (KF) \\ \cline{2-3} 
Model & ADE$\downarrow$              & ADE/FDE$\downarrow$               \\ \hline
FCOS3D~\cite{wang2021fcos3d}    &    1.530              &  3.788/5.162                    \\
Monoloco~\cite{bertoni2019monoloco}   &     1.255            &   2.041/2.728                 \\
MT w/o $L_\text{dir}$(ours)     &  1.202   &    1.915/\textbf{2.514}                   \\
MT (ours)    &       \textbf{1.087}           &     \textbf{1.878}/2.544       \\ \hline
\hline
\end{tabular}
\caption{Replacing the learning-based prediction module by handcrafted KF.}
\vspace{-.7cm}
\label{tab:ablation_module}
\end{table}

\subsection{Run-time efficiency}
Our goal is to implement MT in real-world settings, and we evaluated its efficiency using an RTX 3090 GPU. The main bottleneck of the pipeline lies in the perception stage. We compared the efficiency of detection and BEV localization in \Cref{tab:efficiency}. Overall, detecting keypoints and performing BEV localization are more efficient than the image-based FCOS3D~\cite{wang2021fcos3d}. While MT is slightly slower than Monoloco~\cite{bertoni2019monoloco}, it achieves significantly higher precision. The complete pipeline, which includes OpenPifpaf~\cite{kreiss2021openpifpaf} and MT, takes $88.37$ milliseconds per frame, achieving a frame rate of $11.32$ fps. With its lightweight structure, MT is capable of providing real-time predictions.

\begin{table}[!tbp]
\centering
\begin{tabular}{cc}
\hline
\hline
BEV localization       & Inference time (ms) \\ \hline
FCOS3D~\cite{wang2021fcos3d}                 & 396                 \\
OpenPifpaf~\cite{kreiss2021openpifpaf} + MonoLoco~\cite{bertoni2019monoloco}  & \textbf{81.68}               \\
OpenPifpaf~\cite{kreiss2021openpifpaf} + MT (ours) & 82.50               \\ 
\hline
\hline
\end{tabular}
\caption{Run-time efficiency of different models.}
\vspace{-1.cm}
\label{tab:efficiency}
\end{table}

\subsection{HEADS-UP dataset}

We have demonstrated the efficacy and efficiency of MT on the curated autonomous driving dataset. Now, we implement MT on the HEADS-UP dataset~\cite{VIP2024}. Our primary goal is to accurately predict localizations and trajectories to support downstream tasks for visually impaired individuals, such as collision avoidance. For BEV localization, we compare MT with Monoloco~\cite{bertoni2019monoloco}, which showed strong performance on the NuScenes dataset. For trajectory prediction, we benchmark MT against the KF, a widely used predictor in robotics. \Cref{tab:vis_impaired} presents the quantitative results, showing that MT outperforms other baselines in both BEV localization and trajectory prediction tasks. Learning-based models, such as ST~\cite{saadatnejad2024socialtransmotion} and MT, outperform KF in trajectory prediction. The results highlight that the quality of BEV localization is critical, as shown by the superior performance of KF with MT over ST~\cite{saadatnejad2024socialtransmotion} with Monoloco~\cite{bertoni2019monoloco} due to improved BEV localization. Overall, results on the HEADS-UP dataset are better than those on NuScenes. This can be attributed to the fact that most agents in the HEADS-UP dataset are at closer distances, which is largely due to the more confined environments of the campuses where the data was collected compared to open street environments. Qualitative results are shown in Figure~\ref{fig:quantitative_vip}.

\begin{table}[!tbp]
\centering
\begin{tabular}{ccc}
\hline
\hline
               & BEV localization       & Trajectory prediction \\ \cline{2-3} 
               & ADE$\downarrow$        & ADE/FDE$\downarrow$   \\ \hline
MonoLoco~\cite{bertoni2019monoloco} + KF  & \multirow{2}{*}{0.507} & 1.076/1.566           \\
MonoLoco~\cite{bertoni2019monoloco} + ST~\cite{saadatnejad2024socialtransmotion}  &                        & 1.020/1.518           \\ \hline
MT + KF (ours) & \multirow{2}{*}{\textbf{0.462}} & 0.991/1.453           \\
MT (ours)      &                        & \textbf{0.958/1.405}           \\ 
\hline
\hline
\end{tabular}
\caption{Quantitative results on the HEADS-UP dataset~\cite{VIP2024}.}
\vspace{-1.cm}

\label{tab:vis_impaired}
\end{table}

%% file: sections/conclusions.tex
Removing LiDAR reduces system costs but often degrades BEV localization and trajectory prediction. Our proposed MT bridges that gap by using only a monocular camera, with enhanced precision on both tasks. Additionally, we introduce a directional loss function to enhance prediction quality. We evaluate MT on an autonomous driving dataset and one for visually impaired individuals, demonstrating strong performance in real-world conditions. Our work highlights the critical role of accurate BEV localization in trajectory prediction, a factor often overlooked in conventional studies, underscoring the need to consider real-world settings in trajectory prediction research.

\noindent\textbf{Acknowledgment.} Authors would like to thank Zimin Xia and Reyhaneh Hosseininejad for their insightful comments. This project was funded by Honda R\&D Co., Ltd, and Sportradar.